\documentclass[fleqn,10pt]{wlscirep}
\usepackage[utf8]{inputenc}
\usepackage[T1]{fontenc}
\usepackage{float}
\usepackage{subfig}
\usepackage{caption}
\title{An Improved YOLOv8 Approach for Small Target Detection of Rice Spikelet Flowering in Field Environments}

\author[1]{Beizhang Chen}
\author[1]{Jinming Liang}
\author[1,2,3]{Zheng Xiong}
\author[1,2,3]{Ming Pan}
\author[1,2,3]{Xiangbao Meng}
\author[4]{Qingshan Lin}
\author[4]{Qun Ma}
\author[1,*]{Yingping Zhao}

\affil[1]{Shenzhen Institute of Modern Agricultural Equipment, Shenzhen, 518001, China}
\affil[2]{Guangdong Institute of Modern Agricultural Equipment, Guangzhou, 510630, China}
\affil[3]{Key Laboratory of Modern Agricultural Intelligent Equipment in South China, Ministry of Agriculture and Rural Affairs, Guangzhou, 510630, China}
\affil[4]{Agro-Tech Extension Center of Guangdong Province, Guangzhou, 510520, China}
\affil[*]{zhaoyp@simae.cn}


\begin{abstract}
Accurately detecting the flowering time of rice is crucial for timely pollination in hybrid rice seed production. This not only significantly improves pollination efficiency but also provides an important guarantee for increasing rice yield. However, due to the complexity of field production environments and the characteristics of rice spikelets, such as their small target size and short flowering period, achieving automated and precise recognition of spikelet flowering status in practical applications remains challenging.
To address this, this study proposes a method for recognizing rice spikelet flowering based on an improved YOLOv8 object detection model. First, a Bidirectional Feature Pyramid Network (BiFPN) is introduced to replace the original PANet structure, thereby enhancing the fusion of rice spikelet features and improving the utilization efficiency of multi-scale features. Second, to further enhance small object detection capabilities, this study incorporates a p2 small-object detection head into the model. This detection head employs finer feature mapping, significantly reducing the feature loss issues commonly encountered by standard detection heads in the process of detecting small objects. Additionally, considering the scarcity of publicly available datasets for rice spikelet flowering in field production environments, this study constructs a rice spikelet field dataset using a high-resolution RGB camera and data augmentation techniques, providing reliable data support for model training and testing.
Experiments on the test set demonstrate that the improved YOLOv8s-p2 model achieves an mAP@0.5 of 65.9\%, a precision of 67.6\%, a recall of 61.50\%, and an F1-score of 64.41\%. These results represent improvements of 3.10\%, 8.40\%, 10.80\%, and 9.79\%, respectively, compared to the baseline YOLOv8 model. Moreover, the improved YOLOv8s-p2 model achieves an inference speed of 69 f/s on the test set, fully meeting the requirements of practical application scenarios. The proposed model demonstrates high detection accuracy and speed in rice spikelet flowering detection, providing an effective technical solution for automated and precise monitoring in hybrid rice seed production.
\end{abstract}
\begin{document}

\flushbottom
\maketitle
%
%
\thispagestyle{empty}


\section*{Introduction}
In hybrid rice seed production, precise timing of pollination is crucial for achieving both high yield and superior quality. Unlike many other crops, rice is not strictly self-pollinating, so the success of hybrid seed production relies on effective cross-pollination. Specifically, pollen from the paternal anthers must reach the maternal stigma, ensuring that the resulting seeds inherit the best traits from both parents. This carefully orchestrated exchange of genetic material ultimately leads to hybrid rice varieties with enhanced yield potential and overall quality \cite{tang2012study}. Consequently, accurately identifying the flowering status is essential to synchronize pollen release with stigma receptivity, thereby maximizing seed production efficiency. Conventionally, this has been managed through manual observation, which involves human workers monitoring the flowering stages in the field and initiating pollination at optimal times \cite{duan2011Fast, deng2024Non}. However, this process is labor-intensive, subject to human error, and often impractical for large-scale operations. Consequently, there is a growing need for automated detection methods that can reliably identify the flowering status of rice spikelets to optimize hybrid rice production and reduce labor dependency \cite{zhou2019Automated, hu2021GridFree}.

Rice spikelet flowering detection focuses on identifying specific visual indicators that signify the onset of pollination readiness. Two primary features characterize rice flowering: the opening of spikelet hulls and the extrusion of spikelet anthers \cite{tanaka2014flower}. Spikelet hulls open as a natural precursor to the extrusion of anthers, which release pollen essential for fertilization. Anthers, visible as small structures within the open spikelets, must be extruded to enable cross-pollination. Therefore, capturing these features through imaging and classifying their flowering status are crucial steps for effective hybrid seed production. An automated detection system could revolutionize rice production, providing a solution that is not only efficient but also capable of real-time, large-scale monitoring \cite{zareiforoush2015potential}.

Advancements in image processing and machine learning have greatly enhanced crop phenotyping and monitoring, particularly with the advent of deep learning techniques \cite{murphy2024deep}. Methods such as convolutional neural networks (CNNs) have demonstrated remarkable success in complex image recognition tasks.  Traditional image processing approaches for crop monitoring, such as grayscale thresholding or edge detection, are often limited by environmental factors like lighting, crop density, and image background \cite{Sanaeifar2023advancing}. In contrast, deep learning algorithms can autonomously learn relevant features from labeled image data, achieving greater accuracy and adaptability across diverse conditions.

In recent years, deep learning has demonstrated significant potential in the field of agricultural monitoring, particularly in crop detection tasks. These methods have been widely applied to various crops, gradually addressing key challenges such as the detection of small objects, interference from complex field environments, and real-time processing demands. However, despite these advancements, certain limitations persist in adapting these methods to specific plant components and field environments.

MRF-YOLO \cite{liu2023small} was proposed to detect and count unopened cotton bolls, achieving an average precision of 92.75\% and a mean squared error of 1.06 under field conditions. By incorporating a multi-receptive field extraction module and a small-object detection layer, this method effectively addressed small-object detection challenges in agriculture.
Li et al.\cite{li2023tea} improved YOLOv5 by integrating the Squeeze and Excitation Network and using Hungarian matching and Kalman filtering algorithms for tea bud detection and tracking, achieving an average precision of 91.88\% and high correlation(\(R^2 = 0.98\)) between predicted and manual counts. 
Wang et al. \cite{wang2023multiscale}, based on the RetinaNet model, optimized the feature pyramid structure and introduced the CBAM attention mechanism to achieve accurate recognition of small maize tassels. Experimental results demonstrated that the improved model achieved an average precision of 0.9717 and a recall rate of 0.9036, providing a novel approach for multi-scale maize tassel detection.
These works highlight the effectiveness of deep learning in diverse agricultural applications.

In wheat fields, several methods have been developed to address the challenges of occlusion, lighting, and varying spike shapes. For instance,
Wen et al. \cite{wen2022wheat} introduce SpikeRetinaNet with bidirectional feature pyramid network(BiFPN) and soft non-maximum suppression(Soft-NMS), achieving recognition accuracy of 92.62\%. 
Combining YOLOv5 and DeepMAC, Dandrifosse et al.\cite{dandrifosse2022deep} demonstrated a robust approach for wheat ear detection and segmentation, achieving F1 scores of 0.93 and 0.86, respectively. 
To improve the accuracy of detecting wheat Fusarium Head blight(FHB), Zhang et al.\cite{zhang2023enhancing} introduced the Rotation YOLO Wheat Detection(RYWD) network for detecting wheat heads and the Simple Spatial Attention(SSA) network for extracting features, achieving a predictive accuracy of 94.66\%, providing an accurate methods for wheat disease detection.
Considering the strengths of CNNs and Transformers in wheat spike detection tasks,  Guan et al.\cite{guan2024ctwheatnet} developed CTWheatNet with Quad Attention for improved accuracy and real-time performance in wheat spike deetction.
Li et al.\cite{li2024real} proposed a YOLOv7-based detection and DeepSORT-based tracking approach, achieving a mAP@0.5 of 94.9\% and an average counting accuracy exceeding 97.5\%, with real-time performance at 19.2 FPS. overcome the complexity of natural environments and ensure tracking stability.

Despite the similarities between wheat and rice in spike detection tasks, research on rice remains relatively sparse. Rice, as one of the most significant staple crops, accounting for 8\% of global crop production \cite{fao2023yearbook}, rice detection and monitoring have been less extensively studied compared to other crops.

Wang et al. \cite{wang2022field} proposed a field-scale rice detection and counting algorithm for large-scale paddy field images. Based on YOLOv5x, the method introduced duplicate detection removal method using Inside-Overlapping Box(IOB) and Box-Overlapping Union(BOU). The results demonstrated superior performance in detecting high-density rice spikes and handling occlusions.
To accurately determine the flowering time of rice spikelets, Zhang et al.\cite{zhang2021method} used RGB images and applied the Otsu method on the blue channel to identify spikelet anthers. They employed Faster R-CNN and YOLOv3 to recognize spikelet anthers and opening spikelet hulls. Experimental comparisons showed that Faster R-CNN had higher precision.
Subsequently, the same research group utilized hyperspectral technology combined with machine learning to obtain information on rice spikelet flowering stages \cite{zhang2022detection}. They performed feature extraction using Principal Component Analysis (PCA) and Genetic Algorithm (GA), and constructed detection models such as Random Forest (RF), Support Vector Machine (SVM), Backpropagation (BP) Neural Network, and Convolutional Neural Network (CNN). The results showed that the BP Neural Network based on PCA features was the optimal model for detecting rice spikelet flowering, achieving an accuracy of 96\% to 100\%.

The aforementioned studies have advanced the intelligent development of rice cultivation to some extent, but significant challenges remain. The complexity of field environments, such as small target sizes, variations in lighting conditions, background interference, and occlusion, significantly affects detection accuracy and robustness. Moreover, although existing methods have achieved rice spikelet recognition under laboratory conditions, their application in actual field environments is still limited. Similarly, methods like hyperspectral imaging, while capable of capturing rich spectral information and demonstrating high accuracy under experimental conditions, are constrained by issues such as insufficient real-time performance, high computational complexity, and expensive equipment, limiting their feasibility for large-scale field applications.
To address the aforementioned challenges, this study proposes a deep learning-based method for detecting the flowering status of rice spikelets, enabling real-time recognition of flowering states in field environments. The main contributions are as follows:
\begin{itemize} 
    \item Introduced a Bidirectional Feature Pyramid Network (BiFPN) to enhance the feature fusion of rice spikelets, improving the utilization of multi-scale features. 
    \item Added a p2 small object detection head, which provides finer feature maps compared to the original detection head, reducing feature loss during small object detection and significantly improving the accuracy and robustness of spikelet detection. 
    \item Utilized a low-cost, high-precision RGB camera for data collection, and constructed a rice spikelet flowering dataset for field production environments, providing reliable support for model training and testing. 
    \item The hardware used is cost-effective, and the model inference meets the real-time requirements of practical applications, enabling efficient automated monitoring in large-scale field environments, offering a feasible and practical technological solution for rice production. 
\end{itemize}

\section*{Materials and methods}

\subsection*{Data Acquisition}
\begin{figure}[h]
    \includegraphics[width=\textwidth]{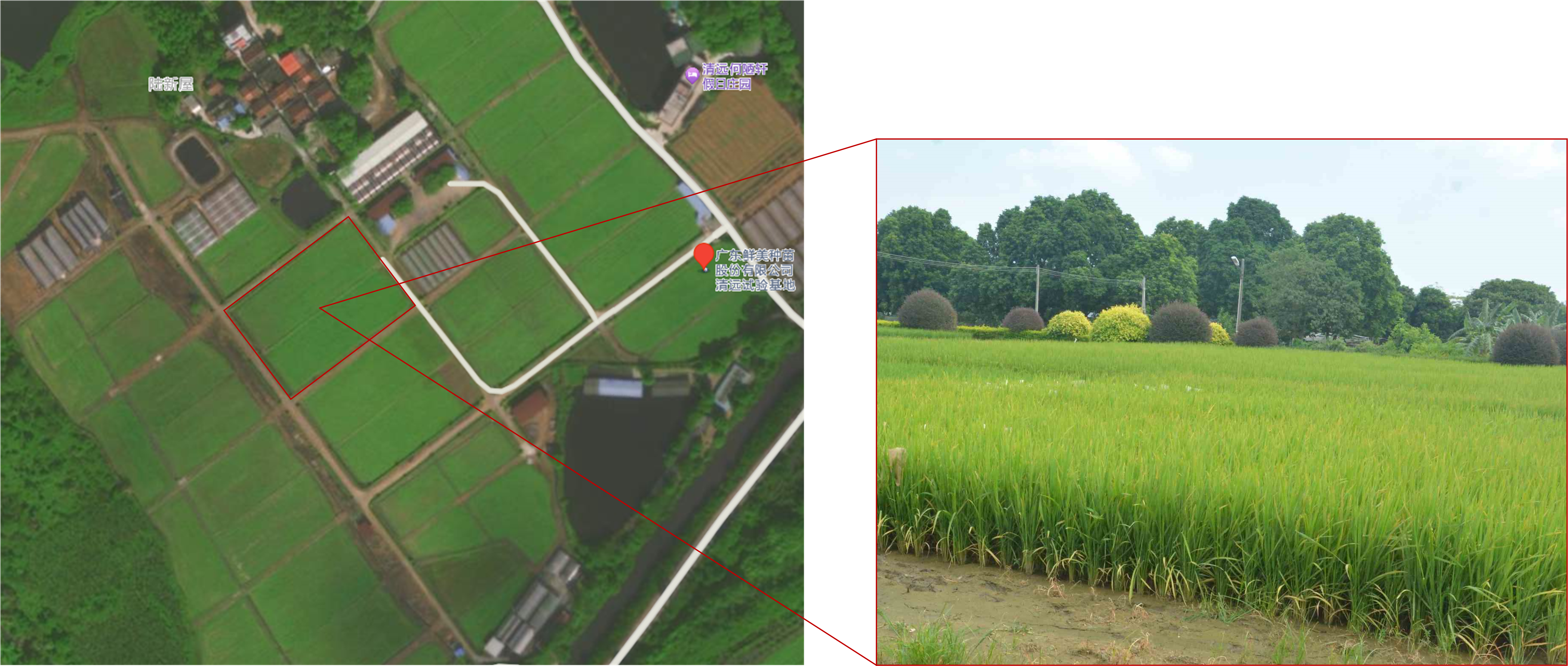}
    \caption[Fig. 1: Data collection site]{Data collection site}
    \label{fig:data_collect}
\end{figure}

The rice spikelet images used in this experiment were collected in early October 2024, between 10:00 AM and 2:00 PM Beijing time, at the rice experimental base in Qingyuan, Guangdong Province (112$^\circ$48.65$'$E, 23$^\circ$38.56$'$N) as shown in Fig. \ref{fig:data_collect}. The images were captured using a Nikon D7100 camera and a smartphone, with resolutions of $4000 \times 6000$ pixels for the camera and $3024 \times 4032$ pixels for the smartphone. During data collection, experimental rice fields in the flowering stage were selected for imaging. 
Two primary imaging methods were employed: 
\begin{itemize}
    \item \textbf{Parallel Imaging}: The lens was kept parallel to the rice spikelets, aiming to capture as many spikelets as possible in a single frame.
    \item \textbf{Overhead Imaging}: The camera was tilted at an angle of $30^\circ$ to $45^\circ$ relative to the horizontal ground, overcoming the occlusion caused by the high planting density of rice to capture more spikelet images.
\end{itemize}

\subsection*{Data Preprocessing}
After data collection, preprocessing and annotation are critical steps that directly affect the accuracy of the rice spikelet flowering detection model. To ensure a diverse and representative dataset, we gather a large number of high-quality images of rice spikelets. During this stage, data cleaning is performed to eliminate blurry or low-quality images, ensuring that the training data remains clear and recognizable. Subsequently, the images are scaled and cropped to emphasize the geometric relationship between the target spikelets and their backgrounds while meeting the input dimension requirements of the model. This step also facilitates accurate and efficient annotation.

\begin{figure}[h]
    \centering
    \includegraphics[width=\linewidth]{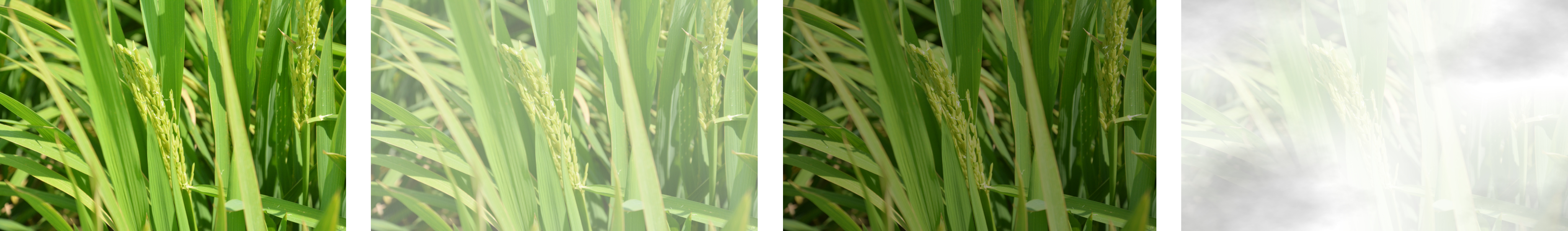}
    \caption{Image augmentation for rice spikelet flowering dataset, from left to right: Original Image, Rain, Lighting, Fog.}
    \label{fig:data-augmentation}
\end{figure}

During preprocessing, various image augmentation techniques are applied, such as rotation, flipping, scaling, and adjustments to brightness and contrast, to systematically expand the standard image dataset for training. Furthermore, to address the challenges posed by environmental factors—such as weather disruptions encountered in agricultural scenarios—we introduce advanced augmentation methods. These include simulating conditions like rain, fog, and varying light intensities to replicate real-world interference. Such enhancements significantly improve the model’s robustness. An example of the effects of these data augmentation techniques is shown in Fig. \ref{fig:data-augmentation}.

\subsection*{Data Annotation and Dataset Splitting}
Data annotation is a critical step in model training. We utilized the professional annotation tool LabelImg to accurately annotate the rice flowering spikelets, completing the annotation process by manually drawing and labeling bounding boxes. Due to the small size of rice spikelets, manual identification is challenging. To ensure standardization and accuracy in the annotations, the results underwent multiple rounds of rigorous review and correction. This allowed for the timely identification and rectification of incorrect labels, significantly enhancing the quality and reliability of the training data. The final dataset consists of 1,584 images, including 4,748 flowering rice spikelet targets.

\begin{figure}[h]
    \centering
    \includegraphics[width=\linewidth]{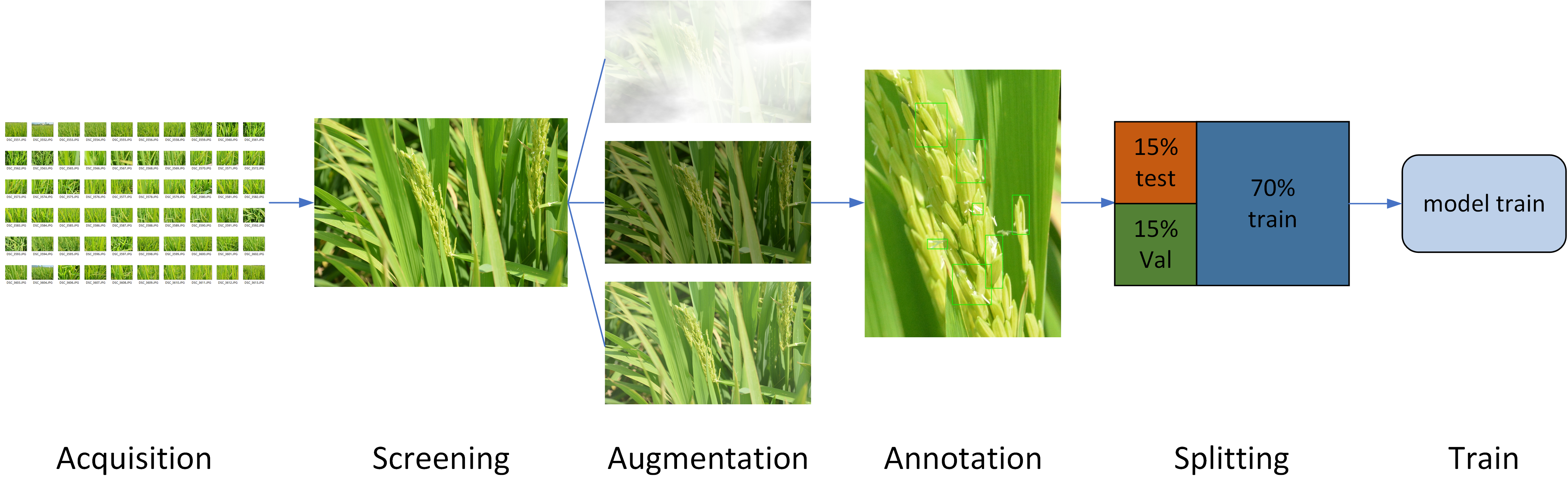}
    \caption{Model training process}
    \label{fig:model-train}
\end{figure}

After completing data annotation, we divided the dataset into test, training, and validation sets in a ratio of 1.5:7:1.5 to improve the model's generalization ability. This split ensures that the model can perform training, validation, and testing on different datasets, thereby enhancing its performance and adaptability to new data. Finally, the model was trained using images augmented through data augmentation, as shown in Fig. \ref{fig:model-train}.

\subsection*{Environment}
The experiments mentioned in this paper were all conducted on a GPU server, with the detailed configuration shown in Table \ref{tab:exp_environment}.

\begin{table}[H]
    \centering
    \begin{tabular}{ll}
        \toprule
        configuration        & parameter                     \\
        \midrule   
        CPU                  &  Intel(R) Xeon(R) Gold 6226R  \\
        GPU                  &  NVIDIA A100 80GB PCIe        \\
        RAM                  &  256GB                        \\
        Operating system     &  Ubuntu 20.04 LTS             \\
        CUDA                 &  11.4                         \\
        PyTorch              &  2.4.0                        \\
    \bottomrule  
    \end{tabular}
    \caption{Experimental environment}
    \label{tab:exp_environment}
\end{table}

\subsection*{Algorithm Evaluation Metrics}
The experiment uses mAP@0.5 Precision, Recall, and \text{F1-score} as evaluation metrics for the algorithm's performance.

mAP: Reflects the average precision of the model for a specific category or query. In this study, we use mAP@0.5, which means the average precision across all categories when the overlap (IoU) between the predicted box and the ground truth box is greater than or equal to 0.5.

\begin{equation}
    \textit{mAP@0.5} = \frac{1}{N} \sum_{i=1}^{N} AP_i
    \label{eq:mAP}
\end{equation}

Precision: Represents the proportion of True Positives among the samples predicted as Positive by the model.

\begin{equation}
    Precision = \frac{TP}{TP + FP}
    \label{eq:Precision}
\end{equation}

Recall: Represents the proportion of all Positives that the model is able to predict.

\begin{equation}
    Recall = \frac{TP}{TP + FN}
    \label{eq:Recall}
\end{equation}

\text{F1-score}: A metric that considers both Precision and Recall, being the harmonic mean of Precision and Recall.

\begin{equation}
    \textit{F1-score} = 2 \times \frac{Precision \times Recall}{Precision + Recall}
    \label{eq:F1-score}
\end{equation}

\section*{Rice Flowering Spikelets Detection Algorithm}
\begin{figure}[H]
    \centering
    \includegraphics[width=0.6\linewidth]{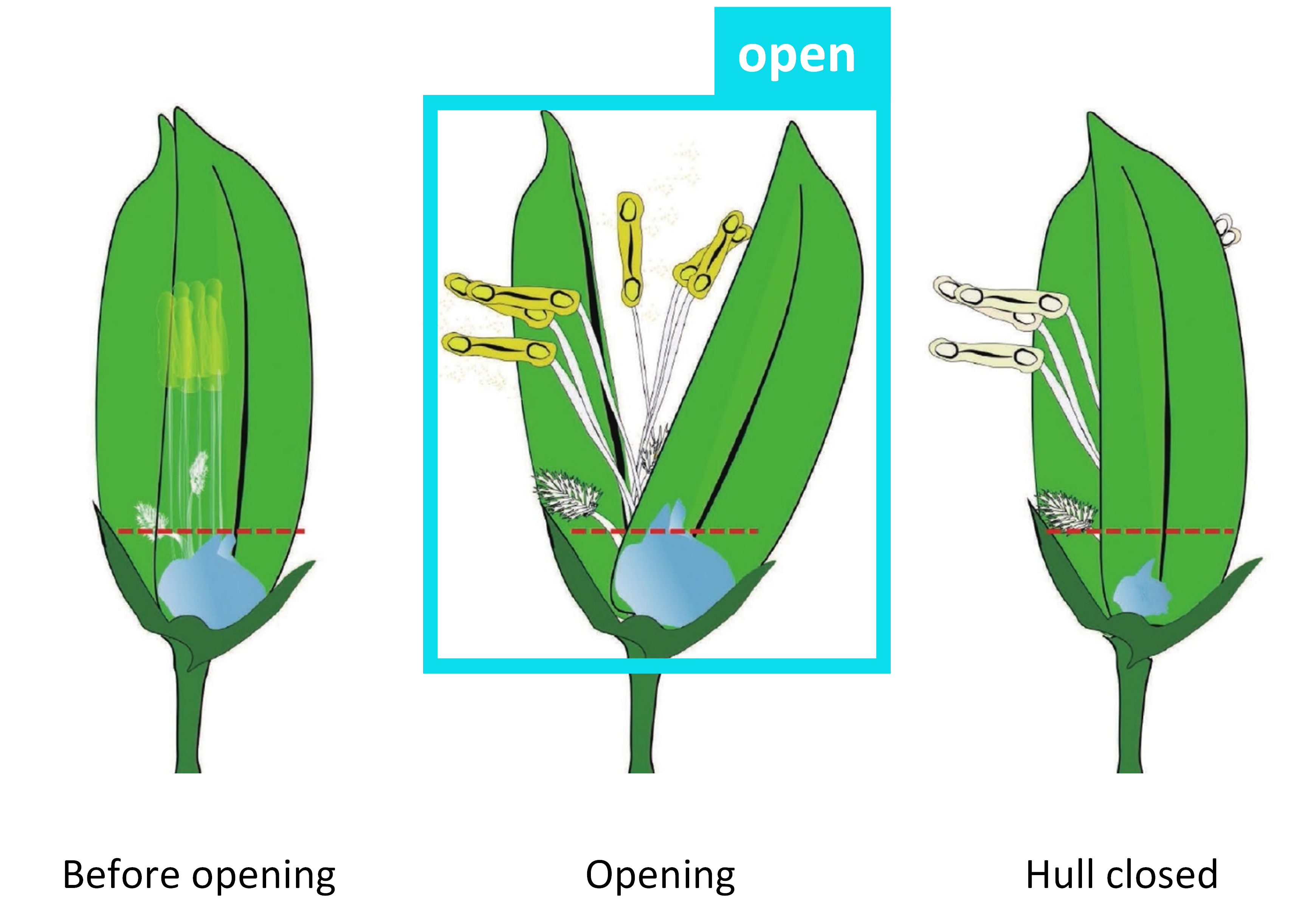}
    \caption{Opening and closure process of rice spikelet}
    \label{fig:rice spikelet opening and colse}
\end{figure}

The rice flowering spikelet detection algorithm aims to accurately identify and classify rice spikelets at different flowering stages.By improving the YOLOv8 model, the algorithm ensures precise localization of open spikelets in dense and complex panicle regions.The different flowering states of rice spikelets are illustrated in Fig. \ref{fig:rice spikelet opening and colse}.

The algorithm primarily utilizes high-resolution RGB images captured from rice fields under varying lighting conditions. These images are collected from different angles to capture diverse perspectives of rice plants. To enhance dataset quality, the locations and labels of flowering spikelets are manually annotated. Additionally, image augmentation techniques are applied to increase dataset diversity and improve the model’s generalization capability.

\begin{figure}[H]
    \centering
    \includegraphics[width=\linewidth]{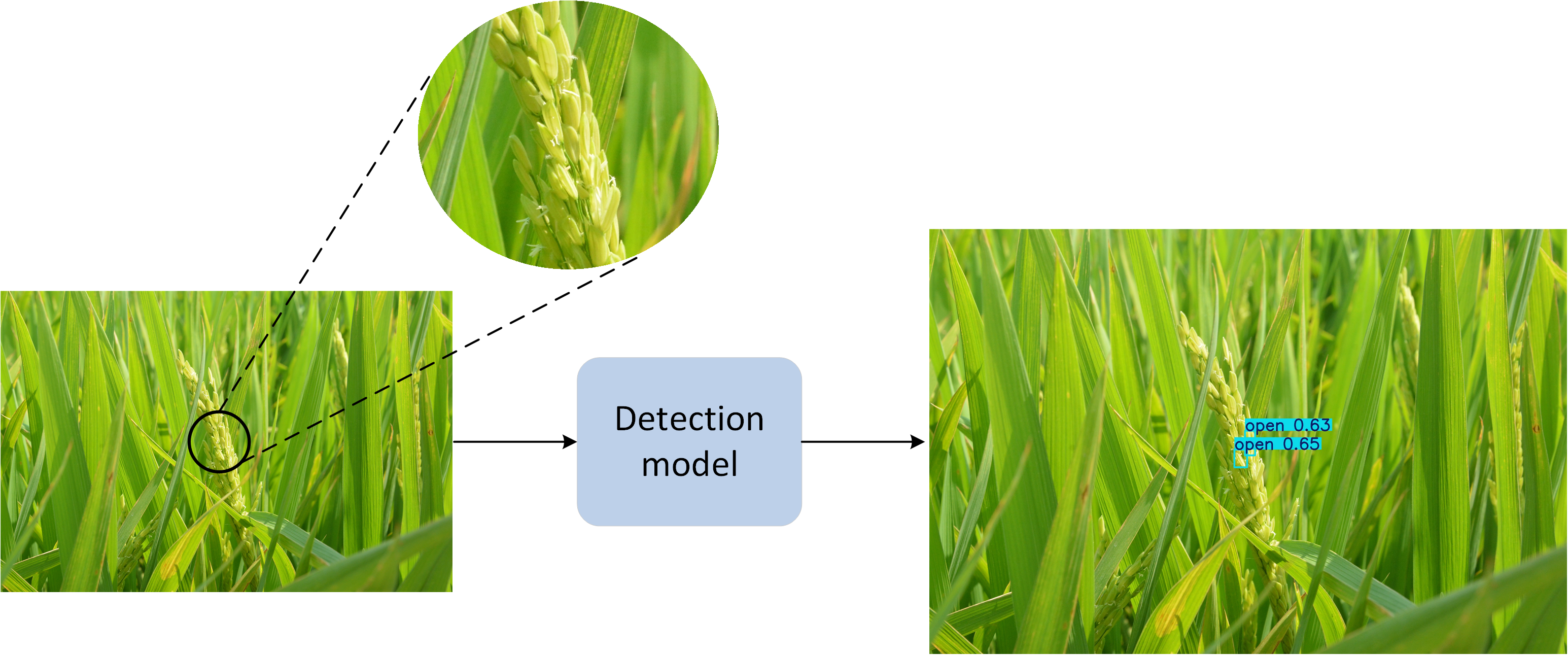}
    \caption{Workflow of the rice spikelet flowering detection algorithm}
    \label{fig:workflow}
\end{figure}

Building upon the strong detection performance of YOLOv8, our model introduces BiFPN (Bidirectional Feature Pyramid Network) and a p2 small-object detection head. These enhancements significantly improve the model’s ability to detect small objects in complex scenes while maintaining real-time inference speed. The workflow of the rice flowering spikelet detection algorithm is illustrated in Fig. \ref{fig:workflow}.

\begin{figure}[h]
    \centering
    \includegraphics[width=\linewidth]{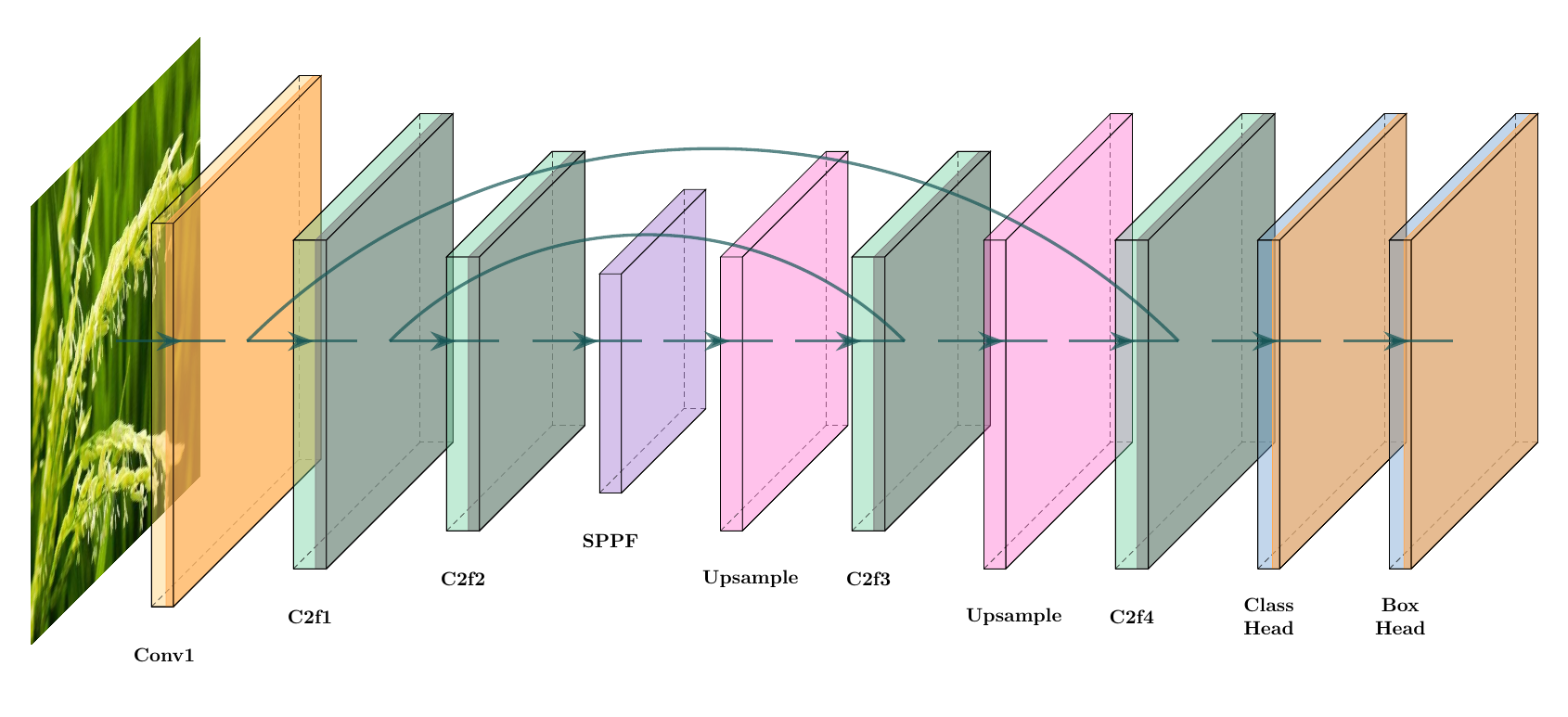}
    \caption{YOLOv8 backbone network}
    \label{fig:yolov8-main}
\end{figure}

\subsection*{Yolov8 Backbone}
\begin{figure}[H]
    \centering
    \includegraphics[width=0.4\linewidth]{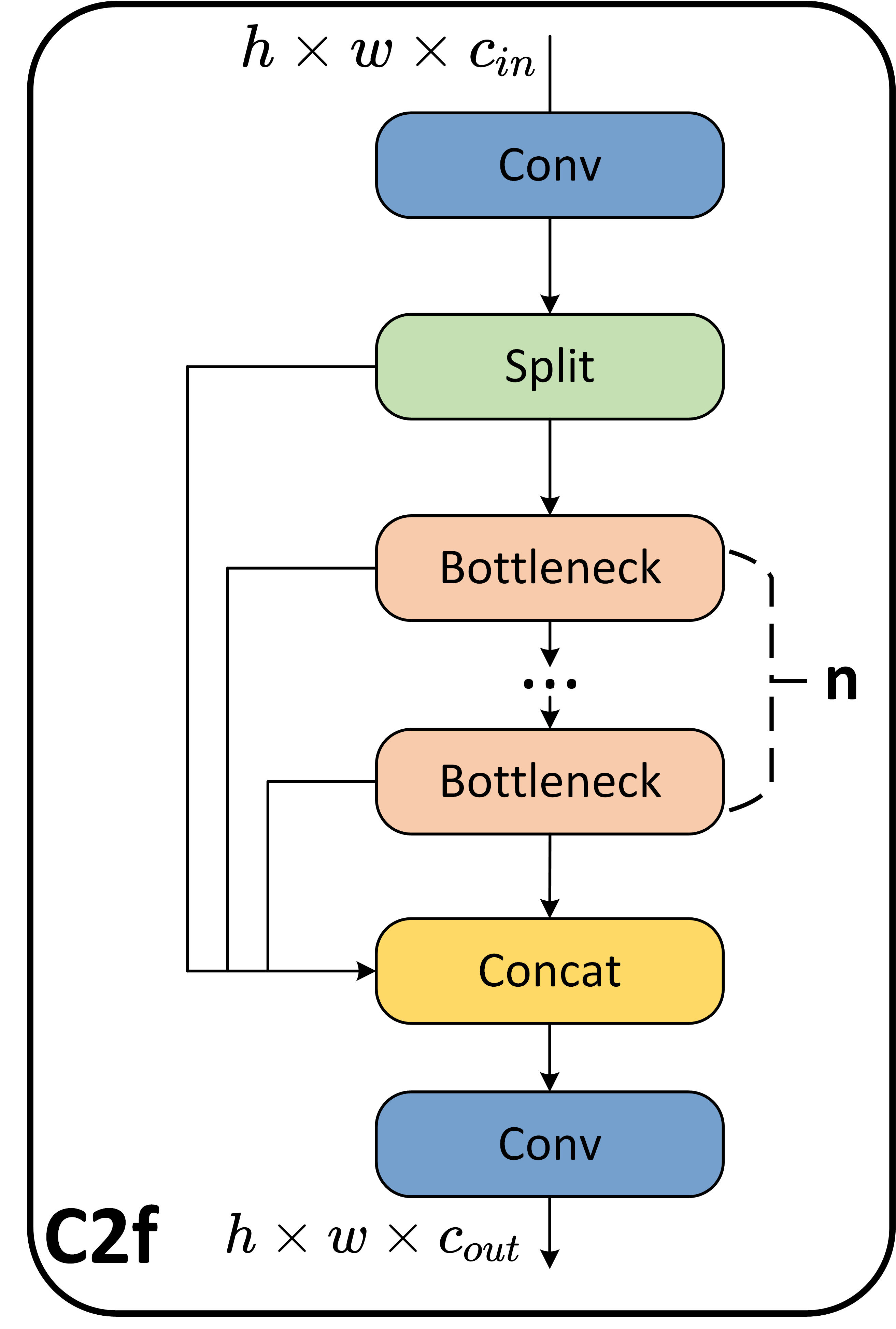}
    \caption{C2f module}
    \label{fig:yolov8-C2f}
\end{figure}

In the YOLOv8 model, the C2f module, derived from the structure of CSPNet, employs a two-branch channel fusion mechanism to achieve cross-stage partial connections through alternating feature paths. This design enhances feature representation capability while effectively reducing computational costs. As shown in Fig. \ref{fig:yolov8-main}, the C2f module strengthens the robustness and detection accuracy of the model by facilitating multi-scale feature fusion. Specifically, the composition of the C2f module, illustrated in Fig. \ref{fig:yolov8-C2f}, offers several advantages: it optimizes features by mixing information from different levels through separated and merged paths, reducing redundant gradients and improving the flow of gradient information. Additionally, it contributes to model lightweighting by enhancing feature propagation through cross-layer connections without significantly increasing computational demands, thereby boosting overall performance. Finally, the diverse model paths in the C2f module expand the effective receptive field, improving the network's ability to detect targets across various scales.

\subsection*{Feature Pyramid}
In order to further enhance the utilization of multi-scale features and improve the efficiency of information flow, we introduced a Bidirectional Feature Pyramid Network (BiFPN) in the Neck component of YOLOv8. The BiFPN not only optimizes feature fusion without significantly increasing computational costs but also improves the precision of multi-scale feature localization for target detection. Compared to the previously utilized PANet structure, BiFPN demonstrates superior performance in multi-scale feature integration, significantly enhancing information flow efficiency and the detection of small targets while maintaining computational efficiency.

\begin{figure}[h]
    \centering
    \includegraphics[width=0.5\linewidth]{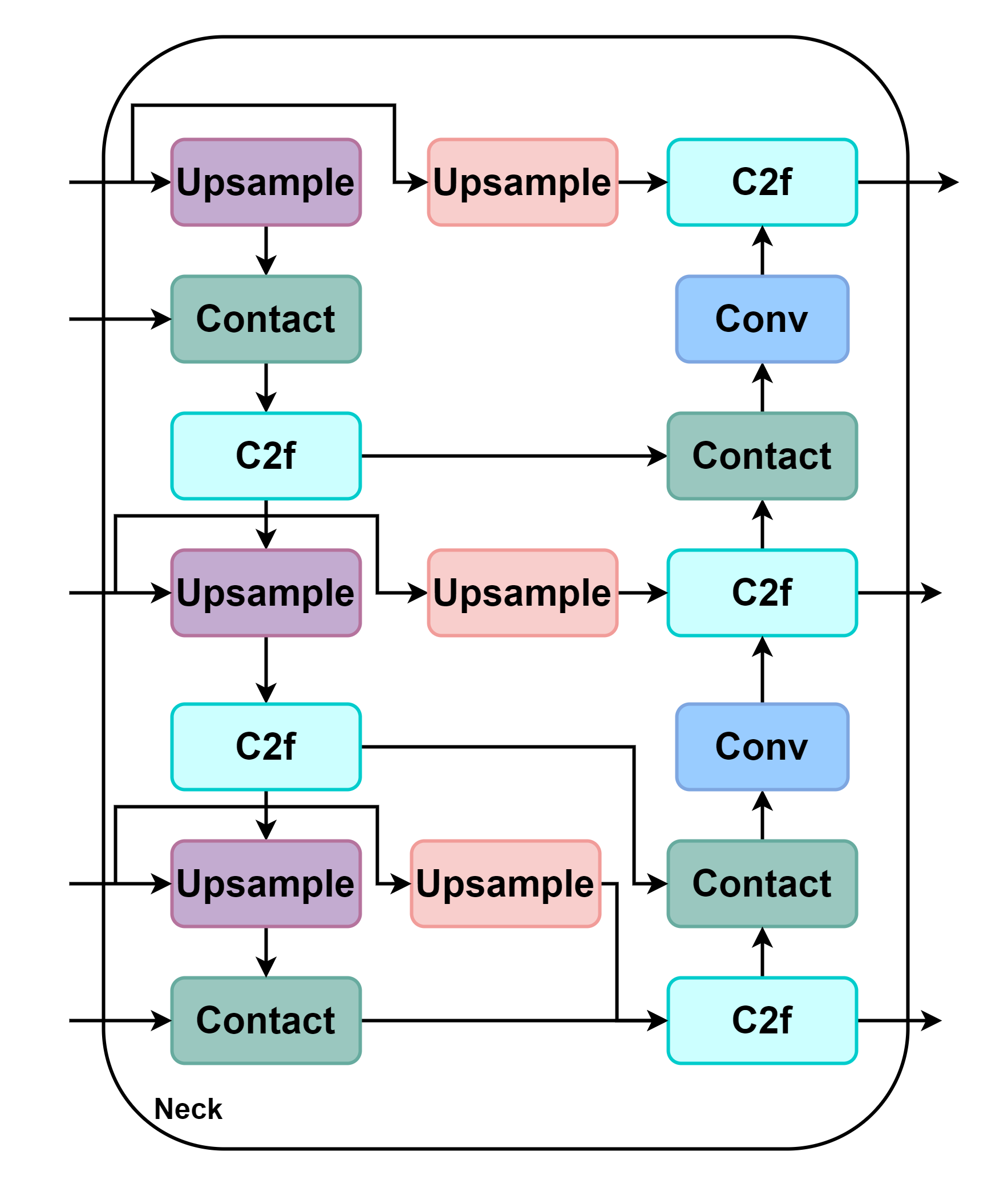}
    \caption{BiFPN network structure}
    \label{fig:BiFPN}
\end{figure}

As illustrated in Fig. \ref{fig:BiFPN}, the BiFPN module introduces several key innovations and advantages. First, enhanced feature fusion is achieved through an efficient bidirectional connection mechanism, enabling features from different scales to communicate and collaborate. This mechanism is complemented by the introduction of weighted sums, which accurately adjust the importance of features at different layers, resulting in more effective feature fusion. Furthermore, optimized information flow is realized by leveraging repeated feature fusion. This process allows features to flow through the network multiple times, enriching contextual information from bottom to top. Such reinforcement significantly improves the network’s ability to perceive small targets and detect objects in complex backgrounds.

In terms of detection performance, the fine integration of features within BiFPN enhances the accuracy and precision of YOLOv8. This improvement is particularly evident in scenarios involving small or densely packed objects, as well as in tasks requiring the detection of targets across varying scales. Additionally, BiFPN offers flexible extensibility, making it suitable for diverse object detection tasks. Its adaptability enables seamless integration into the YOLOv8 framework while maintaining computational efficiency during feature extraction.

Finally, efficient computation is a notable advantage of BiFPN. By employing weighted addition for feature layers and a simplified repeated fusion process, BiFPN optimizes resource usage without significantly increasing computational overhead. This ensures high detection accuracy while preserving the model's speed, a critical factor for real-time detection applications.

With the integration of BiFPN, YOLOv8 demonstrates stronger adaptability to multi-scale and complex scenes, greatly enhancing the reliability and effectiveness of object detection in practical scenarios. This capability is particularly crucial in real-time detection tasks with complex backgrounds, such as rice spikelet flowering detection. By achieving a balance between high detection accuracy and computational efficiency, BiFPN equips YOLOv8 to excel in scenarios with constrained resources and diverse environmental conditions.

\subsection*{P2 Detection Head}
For small target detection, we introduced the p2 small target detection head. By enhancing the resolution capability of low-level features, the p2 detection head extracts feature information of small targets more precisely, allowing for better localization and classification of small targets. This change helps improve the detection accuracy of small targets in complex backgrounds.

The p2 small target detection head is specifically designed for detecting small objects and enhances the network's ability to capture subtle features, contributing to improved detection accuracy. The standard detection head of traditional YOLOv8 may suffer from feature loss in small target scenarios, while the p2 detection head strengthens the identification capability for small-scale targets through finer feature mapping, making it suitable for small and dispersed targets like rice spikelets.

The overall improved YOLOv8 model is better suited to meet the identification needs of small target objects within complex backgrounds during the detection tasks of rice spikelets, enhancing the model's practicality and detection accuracy.

\section*{Results and analysis}
To validate the performance of the YOLOv8s-p2 proposed in this paper for rice spikelet recognition in actual field environments, several widely used models were selected for experiments on our rice spikelet dataset, including FasterRCNN \cite{ren2017fasterrcnn}, SSD \cite{Liu2016ssd}, and EfficientDet \cite{tan2020efficientdet}. In addition, following YOLOv8, YOLOv11 \cite{khanam2024yolov11}, the latest real-time object detection model released by Ultralytics, features fewer model parameters and higher mAP, making it an important comparison object in this study.

\subsection*{Performance Analysis of YOLOv8s in Rice Spikelet Flowering Detection}
To select a suitable object detection model for detecting rice spikelet flowering, we compared the performance of YOLOv8s with several other popular detection models. As shown in Table \ref{tab:Preliminary Comparison}, first, YOLOv8s achieved a mAP@0.5 of 62.80\%, significantly higher than the other models. Compared to Faster-RCNN (52.89\%), SSD (56.59\%), and EfficientDet (59.57\%), YOLOv8s demonstrated stronger object detection capability in the rice spikelet flowering detection task. Secondly, YOLOv8s had a precision of 59.20\%, which is lower than SSD (83.87\%) and EfficientDet (78.00\%), but its recall rate reached 50.70\%, which is significantly higher than SSD (30.79\%) and EfficientDet (41.05\%). This result indicates that YOLOv8s strikes a better balance in reducing false positives and false negatives. In terms of F1-score, YOLOv8s achieved the highest value of 54.62\%, the best among all models, suggesting that YOLOv8s can achieve an optimal trade-off between precision and recall, demonstrating more stable and reliable overall detection performance. Finally, detection speed is a crucial metric for real-time detection tasks, and YOLOv8s achieved a FPS of 109, far surpassing the other models (Faster-RCNN 9 FPS, SSD 31 FPS, EfficientDet 8 FPS). This result shows that YOLOv8s is capable of meeting the needs for real-time field detection and efficient data collection.
\begin{table}[H]
    \centering
        \begin{tabular}{lccccc}
            \toprule
            model        & mAP@0.5(\%)    & Precision(\%) & Recall(\%) & \text{F1-score}(\%) & FPS(f/s) \\
            \midrule
            Faster-RCNN  & 52.89\%        & 46.85\%       & 61.84\%    &  53.31\%            & 9        \\
            SSD          & 56.59\%        & 83.87\%       & 30.79\%    &  45.04\%            & 31       \\
            EfficientDet & 59.57\%        & 78.00\%       & 41.05\%    &  53.79\%            & 8        \\
            YOLOv8s      & 62.80\%        & 59.20\%       & 50.70\%    &  54.62\%            & 109      \\     
            \bottomrule
        \end{tabular}
        \caption{Preliminary comparison of the performance of different detection models for rice spikelet flowering detection}
        \label{tab:Preliminary Comparison}
\end{table}

\subsection*{Performance Analysis of Improved YOLOv8s-p2 for Rice Spikelet Detection}

\begin{figure}[H]
\centering
    \subfloat[\centering ground truth]{\includegraphics[width=0.5\textwidth]{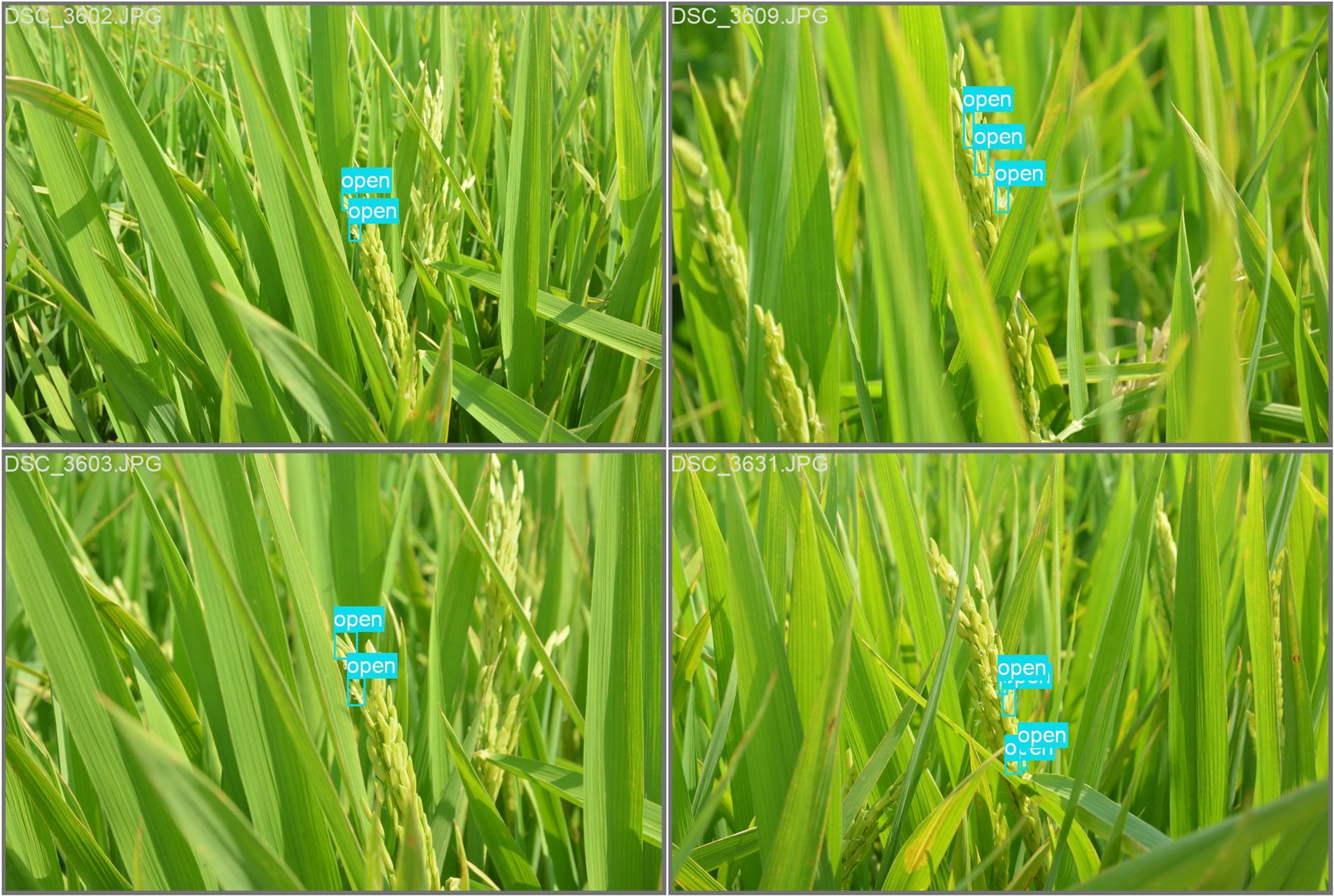}}\\
    \subfloat[\centering YOLOv8]{\includegraphics[width=0.5\textwidth]{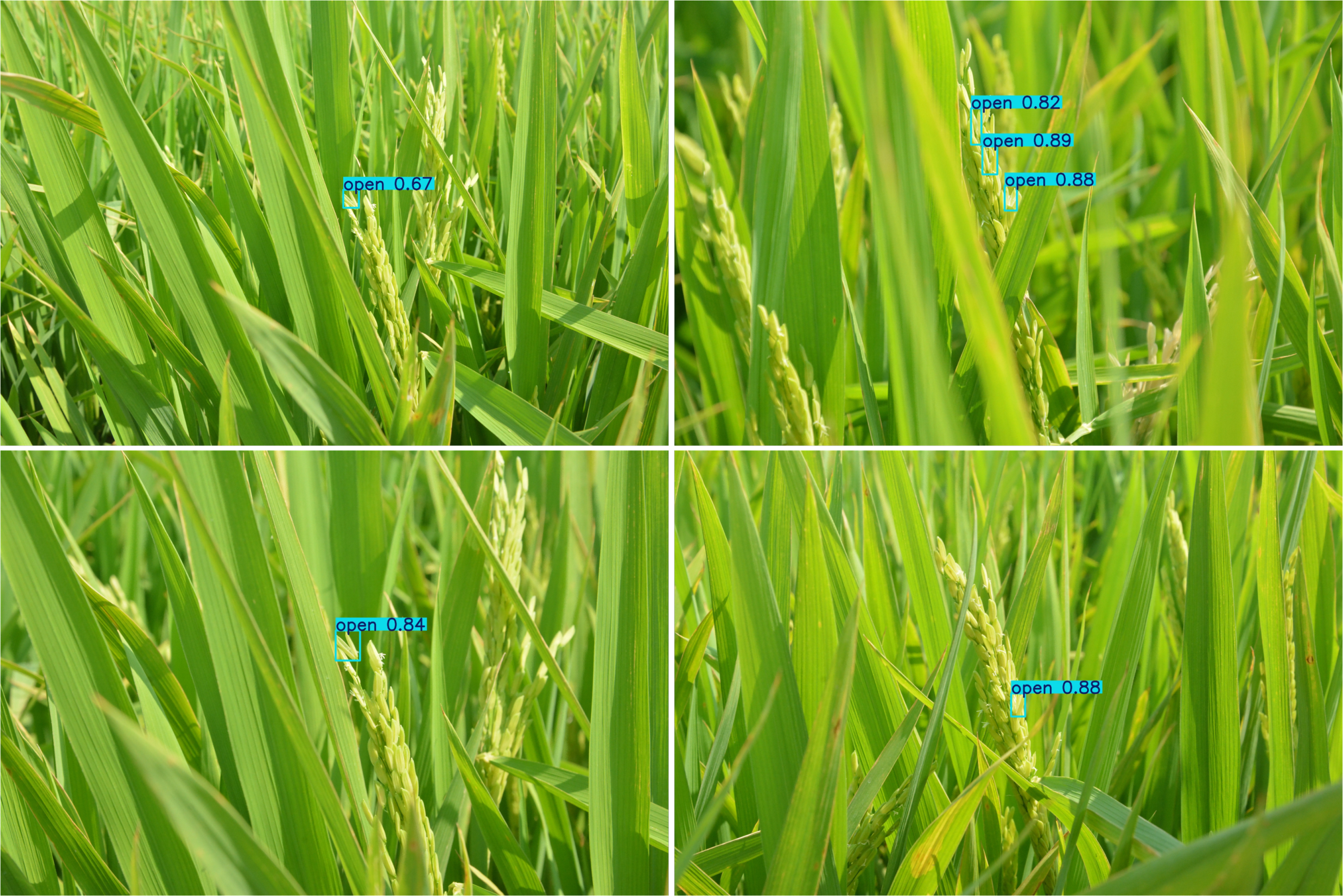}}\\
    \subfloat[\centering YOLOv8s-p2]{\includegraphics[width=0.5\textwidth]{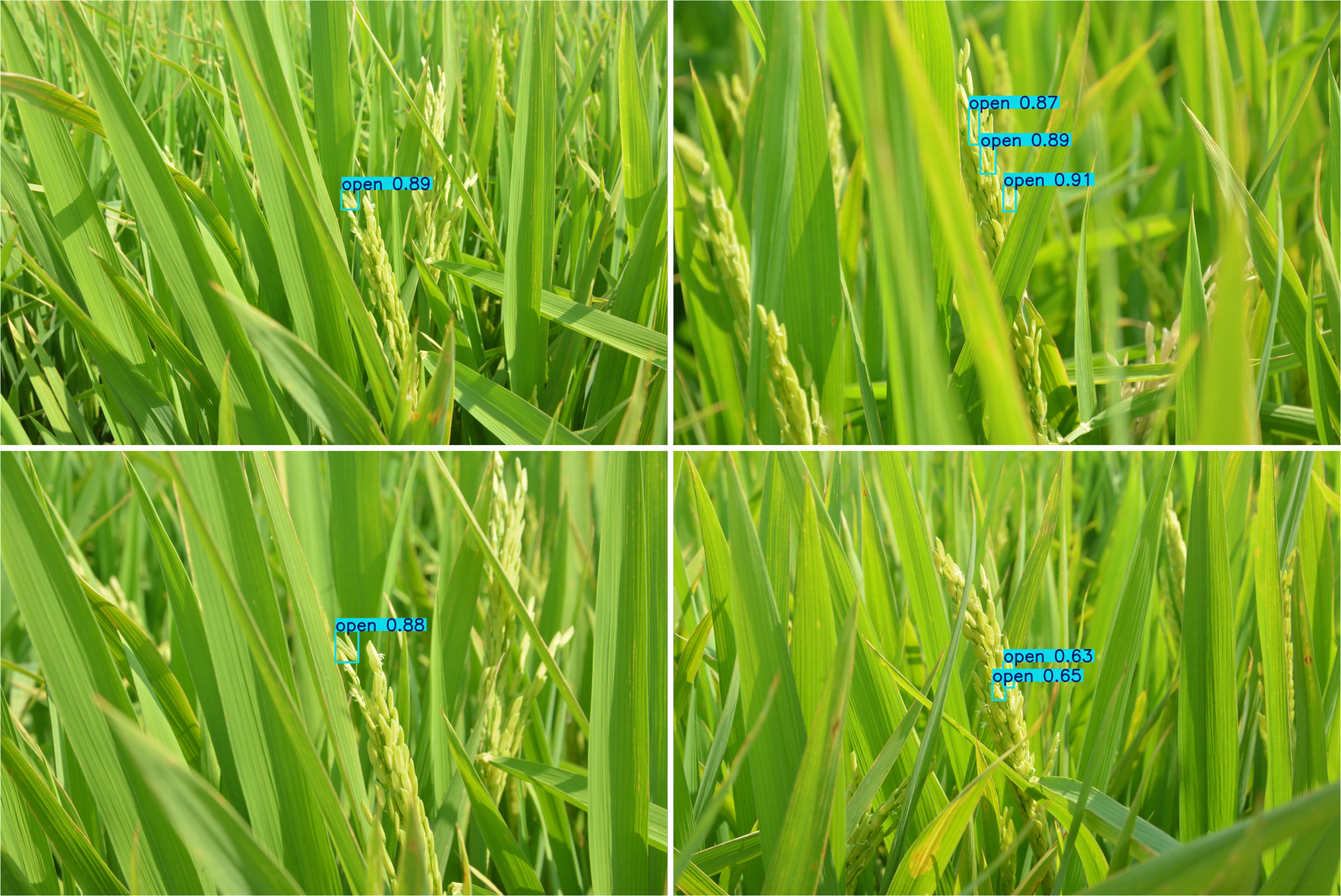}}
    \caption{Comparison of detection between ground truth, YOLOv8, and YOLOv8s-p2 on rice spikelet images. (a) Detection ground truth; (b) Detection effect of YOLOv8; (c) Detection effect of YOLOv8s-p2.}
    \label{fig:detect effect}
\end{figure}

In the rice spikelet flowering detection task, the detection result of ground truth, yolov8 and yolov8-p2 are shown in Fig. \ref{fig:detect effect}, and the experimental results in Table \ref{tab:flowering detection} demonstrate that the improved YOLOv8s-p2 model exhibits a significant performance improvement over the baseline YOLOv8s model. Specifically, YOLOv8s-p2 achieved a mAP@0.5 of 65.90\%, a precision of 67.60\%, a recall of 61.50\%, and an F1-score of 64.41\%, representing improvements of 3.10\%, 8.40\%, 10.80\%, and 9.79\%, respectively, compared to the baseline model. This indicates that YOLOv8s-p2 has higher detection accuracy and stronger small-object detection capability in complex backgrounds. Although its detection speed (69 FPS) is slightly lower than the baseline model (109 FPS), it still fully meets the real-time requirements of the task. Compared to YOLOv11, while YOLOv8s-p2 slightly lags behind in mAP@0.5, its precision, recall, and F1-score have increased by 0.70\%, 2.50\%, and 1.71\%, respectively, showing better balance in reducing false positives and false negatives, and demonstrating higher detection quality. Although YOLOv11 boasts an ultra-high FPS of 217, its slightly lower F1-score may affect the detection reliability for tasks. Overall, YOLOv8s-p2 exhibits stronger scene adaptability and balanced performance in rice spikelet flowering detection, making it a more optimal solution.

\begin{table}[h]
    \centering
        \begin{tabular}{lccccc}
            \toprule
            model        & mAP@0.5(\%)    & Precision(\%) & Recall(\%) & \text{F1-score}(\%) & FPS(f/s) \\
            \midrule
            YOLOv8s      & 62.80\%        & 59.20\%       & 50.70\%    &  54.62\%            &  109     \\ 
            YOLOv11      & 66.40\%        & 66.90\%       & 59.00\%    &  62.70\%            &  217      \\
            YOLOv8s-p2   & 65.90\%        & 67.60\%       & 61.50\%    &  64.41\%            &  69      \\
            \bottomrule
        \end{tabular}
    \caption{Comparison of the performance of improved model for rice spikelet flowering detection}
    \label{tab:flowering detection}
\end{table}

\section*{Discussion}
In this study, the proposed YOLOv8s-p2 model demonstrates excellent performance in the task of rice spikelet flowering. Compared to the baseline model YOLOv8s, YOLOv8s-p2 achieves significant improvements across several key performance metrics, especially in precision. This indicates that the model has better small object detection capability and overall detection quality, particularly in complex environments.

Reviewing the existing literature on object detection in agricultural applications, YOLO models have been widely applied in crop detection tasks and are generally considered efficient solutions in the field of object detection. YOLO models, especially in real-time detection tasks that require high accuracy, show clear advantages. Our experimental results further confirm this, as YOLOv8s-p2 not only improves detection accuracy compared to other object detection models (such as Faster-RCNN, SSD, and EfficientDet), but also achieves a better balance in reducing false positives and false negatives, thereby enhancing the model's robustness in complex backgrounds. This stands in contrast to the challenges faced by traditional machine learning methods in balancing speed and accuracy for specific tasks, further proving the unique advantages of YOLOv8s-p2 in agricultural detection tasks.

Compared to the baseline model, the improved YOLOv8s-p2 achieves significant gains in both mAP and F1-score, while maintaining high detection speed. Therefore, YOLOv8s-p2 strikes a better balance between real-time detection and high-accuracy detection, making it particularly suitable for agricultural tasks like rice spikelet flowering detection.

Building on the high real-time performance of YOLO models, future work will focus on deploying cameras and rail systems in rice production environments to capture video streams for real-time counting of rice spikelet flowering. In this context, subsequent research will explore methods for avoiding duplicate target counts between consecutive frames in moving video streams to improve accuracy and efficiency. However, small object detection still faces significant challenges, particularly in complex environments where small targets, cluttered backgrounds, lighting changes, and occlusions can impact detection performance. Therefore, future research could further optimize the YOLOv8s-p2 model to improve detection accuracy, particularly in challenging real-world conditions such as lighting variations, occlusions, and complex backgrounds.

Furthermore, considering the need for deployment on edge devices, an important direction for future research will be to compress the model and optimize inference speed, enabling efficient real-time detection on resource-constrained devices. Solving this issue will help enhance the applicability of the YOLOv8s-p2 model in rice production, driving further advancements in the automation and intelligence of rice farming.

\section*{Conclusion}
Accurate detection of rice spikelet flowering is essential for understanding important physiological traits such as pollination efficiency and grain setting rate. It is crucial for improving hybrid rice seed production efficiency and ensuring the stability of rice production. To address the issue of scarce rice spikelet flowering data, we have constructed a dataset consisting of 1,584 high-resolution images, containing a total of 4,748 spikelet flowering labels. To enhance the accuracy of the detection model, we optimized the YOLOv8s model by incorporating the BiFPN feature fusion network to strengthen the feature fusion of rice spikelets and adding a p2 small-object detection head to improve its ability to detect small targets. Experimental results show that, compared to YOLOv8s, the improved YOLOv8s-p2 model achieves higher accuracy and recall rates for detecting rice spikelet flowering in complex backgrounds, demonstrating better overall performance and providing reliable technical support for the precise detection of rice spikelet flowering status. In the future, this model can be further applied in practical agricultural production scenarios, offering strong support for improving hybrid rice seed production efficiency, optimizing breeding strategies, and ensuring the stability of rice yields.

\bibliography{sample}

\begin{thebibliography}{10}
\urlstyle{rm}
\expandafter\ifx\csname url\endcsname\relax
  \def\url#1{\texttt{#1}}\fi
\expandafter\ifx\csname urlprefix\endcsname\relax\def\urlprefix{URL }\fi
\expandafter\ifx\csname doiprefix\endcsname\relax\def\doiprefix{DOI: }\fi
\providecommand{\bibinfo}[2]{#2}
\providecommand{\eprint}[2][]{\url{#2}}

\bibitem{tang2012study}
\bibinfo{author}{Tang, C.~Z.} \emph{et~al.}
\newblock \bibinfo{journal}{\bibinfo{title}{Study status and developmental strategies of mechanical pollination for hybrid rice breeding}}.
\newblock {\emph{\JournalTitle{Transactions of the Chinese Society of Agricultural Engineering}}} \textbf{\bibinfo{volume}{28}}, \bibinfo{pages}{1--7} (\bibinfo{year}{2012}).

\bibitem{duan2011Fast}
\bibinfo{author}{Duan, L.} \emph{et~al.}
\newblock \bibinfo{journal}{\bibinfo{title}{Fast discrimination and counting of filled/unfilled rice spikelets based on bi-modal imaging}}.
\newblock {\emph{\JournalTitle{Computers and Electronics in Agriculture}}} \textbf{\bibinfo{volume}{75}}, \bibinfo{pages}{196--203}, \doiprefix\url{10.1016/j.compag.2010.11.004} (\bibinfo{year}{2011}).

\bibitem{deng2024Non}
\bibinfo{author}{Deng, R.} \emph{et~al.}
\newblock \bibinfo{journal}{\bibinfo{title}{Non-destructive measurement of rice spikelet size based on panicle structure using deep learning method}}.
\newblock {\emph{\JournalTitle{Agronomy}}} \textbf{\bibinfo{volume}{14}}, \bibinfo{pages}{2398}, \doiprefix\url{10.3390/agronomy14102398} (\bibinfo{year}{2024}).

\bibitem{zhou2019Automated}
\bibinfo{author}{Zhou, C.} \emph{et~al.}
\newblock \bibinfo{journal}{\bibinfo{title}{Automated counting of rice panicle by applying deep learning model to images from unmanned aerial vehicle platform}}.
\newblock {\emph{\JournalTitle{Sensors}}} \textbf{\bibinfo{volume}{19}}, \bibinfo{pages}{3106}, \doiprefix\url{10.3390/s19143106} (\bibinfo{year}{2019}).

\bibitem{hu2021GridFree}
\bibinfo{author}{Hu, Y.} \& \bibinfo{author}{Zhang, Z.}
\newblock \bibinfo{journal}{\bibinfo{title}{Gridfree: A python package for interactive grain counting and measuring}}.
\newblock {\emph{\JournalTitle{Plant Physiology}}} \textbf{\bibinfo{volume}{186}}, \bibinfo{pages}{2239--2252}, \doiprefix\url{10.1093/plphys/kiab226} (\bibinfo{year}{2021}).

\bibitem{tanaka2014flower}
\bibinfo{author}{Tanaka, W.}, \bibinfo{author}{Toriba, T.} \& \bibinfo{author}{Hirano, H.-Y.}
\newblock \bibinfo{title}{Chapter eight - flower development in rice}.
\newblock In \bibinfo{editor}{Fornara, F.} (ed.) \emph{\bibinfo{booktitle}{The Molecular Genetics of Floral Transition and Flower Development}}, vol.~\bibinfo{volume}{72} of \emph{\bibinfo{series}{Advances in Botanical Research}}, \bibinfo{pages}{221--262}, \doiprefix\url{10.1016/B978-0-12-417162-6.00008-0} (\bibinfo{publisher}{Elsevier}, \bibinfo{year}{2014}).

\bibitem{zareiforoush2015potential}
\bibinfo{author}{Zareiforoush, H.}, \bibinfo{author}{Minaei, S.}, \bibinfo{author}{Alizadeh, M.~R.} \& \bibinfo{author}{Banakar, A.}
\newblock \bibinfo{journal}{\bibinfo{title}{Potential applications of computer vision in quality inspection of rice: A review}}.
\newblock {\emph{\JournalTitle{Food Engineering Reviews}}} \textbf{\bibinfo{volume}{7}}, \bibinfo{pages}{321--345}, \doiprefix\url{10.1007/s12393-014-9101-z} (\bibinfo{year}{2015}).

\bibitem{murphy2024deep}
\bibinfo{author}{Murphy, K.~M.}, \bibinfo{author}{Ludwig, E.}, \bibinfo{author}{Gutierrez, J.} \& \bibinfo{author}{Gehan, M.~A.}
\newblock \bibinfo{journal}{\bibinfo{title}{Deep learning in image-based plant phenotyping}}.
\newblock {\emph{\JournalTitle{Annual Review of Plant Biology}}} \textbf{\bibinfo{volume}{75}}, \bibinfo{pages}{1--20}, \doiprefix\url{10.1146/annurev-arplant-070523-042828} (\bibinfo{year}{2024}).

\bibitem{Sanaeifar2023advancing}
\bibinfo{author}{Sanaeifar, A.} \emph{et~al.}
\newblock \bibinfo{journal}{\bibinfo{title}{Advancing precision agriculture: The potential of deep learning for cereal plant head detection}}.
\newblock {\emph{\JournalTitle{Computers and Electronics in Agriculture}}} \textbf{\bibinfo{volume}{209}}, \bibinfo{pages}{107875}, \doiprefix\url{10.1016/j.compag.2023.107875} (\bibinfo{year}{2023}).

\bibitem{liu2023small}
\bibinfo{author}{Liu, Q.}, \bibinfo{author}{Zhang, Y.} \& \bibinfo{author}{Yang, G.}
\newblock \bibinfo{journal}{\bibinfo{title}{Small unopened cotton boll counting by detection with mrf-yolo in the wild}}.
\newblock {\emph{\JournalTitle{Computers and Electronics in Agriculture}}} \textbf{\bibinfo{volume}{204}}, \bibinfo{pages}{107576}, \doiprefix\url{10.1016/j.compag.2022.107576} (\bibinfo{year}{2023}).

\bibitem{li2023tea}
\bibinfo{author}{Li, Y.}, \bibinfo{author}{Ma, R.}, \bibinfo{author}{Zhang, R.}, \bibinfo{author}{Cheng, Y.} \& \bibinfo{author}{Dong, C.}
\newblock \bibinfo{journal}{\bibinfo{title}{A tea buds counting method based on yolov5 and kalman filter tracking algorithm}}.
\newblock {\emph{\JournalTitle{Plant Phenomics}}} \textbf{\bibinfo{volume}{5}}, \bibinfo{pages}{0030}, \doiprefix\url{10.34133/plantphenomics.0030} (\bibinfo{year}{2023}).

\bibitem{wang2023multiscale}
\bibinfo{author}{Wang, B.} \emph{et~al.}
\newblock \bibinfo{journal}{\bibinfo{title}{Multiscale maize tassel identification based on improved retinanet model and uav images}}.
\newblock {\emph{\JournalTitle{Remote Sensing}}} \textbf{\bibinfo{volume}{15}}, \bibinfo{pages}{2530}, \doiprefix\url{10.3390/rs15102530} (\bibinfo{year}{2023}).

\bibitem{wen2022wheat}
\bibinfo{author}{Wen, C.} \emph{et~al.}
\newblock \bibinfo{journal}{\bibinfo{title}{Wheat spike detection and counting in the field based on spikeretinanet}}.
\newblock {\emph{\JournalTitle{Frontiers in Plant Science}}} \textbf{\bibinfo{volume}{13}}, \bibinfo{pages}{821717}, \doiprefix\url{10.3389/fpls.2022.821717} (\bibinfo{year}{2022}).

\bibitem{dandrifosse2022deep}
\bibinfo{author}{Dandrifosse, S.} \emph{et~al.}
\newblock \bibinfo{journal}{\bibinfo{title}{Deep learning for wheat ear segmentation and ear density measurement: From heading to maturity}}.
\newblock {\emph{\JournalTitle{Computers and Electronics in Agriculture}}} \textbf{\bibinfo{volume}{199}}, \bibinfo{pages}{107161}, \doiprefix\url{10.1016/j.compag.2022.107161} (\bibinfo{year}{2022}).

\bibitem{zhang2023enhancing}
\bibinfo{author}{Zhang, D.-Y.} \emph{et~al.}
\newblock \bibinfo{journal}{\bibinfo{title}{Enhancing wheat fusarium head blight detection using rotation yolo wheat detection network and simple spatial attention network}}.
\newblock {\emph{\JournalTitle{Computers and Electronics in Agriculture}}} \textbf{\bibinfo{volume}{211}}, \bibinfo{pages}{107968}, \doiprefix\url{10.1016/j.compag.2023.107968} (\bibinfo{year}{2023}).

\bibitem{guan2024ctwheatnet}
\bibinfo{author}{Guan, Y.} \emph{et~al.}
\newblock \bibinfo{journal}{\bibinfo{title}{Ctwheatnet: Accurate detection model of wheat ears in field}}.
\newblock {\emph{\JournalTitle{Computers and Electronics in Agriculture}}} \textbf{\bibinfo{volume}{225}}, \bibinfo{pages}{109272}, \doiprefix\url{10.1016/j.compag.2024.109272} (\bibinfo{year}{2024}).

\bibitem{li2024real}
\bibinfo{author}{Li, Z.} \emph{et~al.}
\newblock \bibinfo{journal}{\bibinfo{title}{Real-time detection and counting of wheat ears based on improved yolov7}}.
\newblock {\emph{\JournalTitle{Computers and Electronics in Agriculture}}} \textbf{\bibinfo{volume}{218}}, \bibinfo{pages}{108670}, \doiprefix\url{10.1016/j.compag.2024.108670} (\bibinfo{year}{2024}).

\bibitem{fao2023yearbook}
\bibinfo{author}{FAO}.
\newblock \emph{\bibinfo{title}{{World Food and Agriculture – Statistical Yearbook 2023}}} (\bibinfo{publisher}{FAO}, \bibinfo{address}{Rome, Italy}, \bibinfo{year}{2023}).

\bibitem{wang2022field}
\bibinfo{author}{Wang, X.} \emph{et~al.}
\newblock \bibinfo{journal}{\bibinfo{title}{Field rice panicle detection and counting based on deep learning}}.
\newblock {\emph{\JournalTitle{Frontiers in Plant Science}}} \textbf{\bibinfo{volume}{13}}, \bibinfo{pages}{966495}, \doiprefix\url{10.3389/fpls.2022.966495} (\bibinfo{year}{2022}).

\bibitem{zhang2021method}
\bibinfo{author}{Zhang, Y.} \emph{et~al.}
\newblock \bibinfo{journal}{\bibinfo{title}{Method for detecting rice flowering spikelets using visible light images}}.
\newblock {\emph{\JournalTitle{Transactions of the Chinese Society of Agricultural Engineering}}} \textbf{\bibinfo{volume}{37}}, \bibinfo{pages}{253--262} (\bibinfo{year}{2021}).

\bibitem{zhang2022detection}
\bibinfo{author}{Zhang, Y.} \emph{et~al.}
\newblock \bibinfo{journal}{\bibinfo{title}{Detection of rice spikelet flowering for hybrid rice seed production using hyperspectral technique and machine learning}}.
\newblock {\emph{\JournalTitle{Agriculture}}} \textbf{\bibinfo{volume}{12}}, \bibinfo{pages}{755}, \doiprefix\url{10.3390/agriculture12060755} (\bibinfo{year}{2022}).

\bibitem{ren2017fasterrcnn}
\bibinfo{author}{Ren, S.}, \bibinfo{author}{He, K.}, \bibinfo{author}{Girshick, R.} \& \bibinfo{author}{Sun, J.}
\newblock \bibinfo{journal}{\bibinfo{title}{Faster r-cnn: Towards real-time object detection with region proposal networks}}.
\newblock {\emph{\JournalTitle{IEEE Transactions on Pattern Analysis and Machine Intelligence}}} \textbf{\bibinfo{volume}{39}}, \bibinfo{pages}{1137--1149}, \doiprefix\url{10.1109/TPAMI.2016.2577031} (\bibinfo{year}{2017}).

\bibitem{Liu2016ssd}
\bibinfo{author}{Liu, W.} \emph{et~al.}
\newblock \bibinfo{title}{Ssd: Single shot multibox detector}.
\newblock In \emph{\bibinfo{booktitle}{Computer Vision—ECCV 2016}}, \bibinfo{pages}{21--37}, \doiprefix\url{10.1007/978-3-319-46448-0_2} (\bibinfo{publisher}{Springer International Publishing}, \bibinfo{address}{Cham}, \bibinfo{year}{2016}).

\bibitem{tan2020efficientdet}
\bibinfo{author}{Tan, M.}, \bibinfo{author}{Pang, R.} \& \bibinfo{author}{Le, Q.}
\newblock \bibinfo{title}{Efficientdet: Scalable and efficient object detection}.
\newblock In \emph{\bibinfo{booktitle}{Proceedings of the IEEE/CVF Conference on Computer Vision and Pattern Recognition}}, \bibinfo{pages}{10781--10790}, \doiprefix\url{10.1109/CVPR42600.2020.01079} (\bibinfo{year}{2020}).

\bibitem{khanam2024yolov11}
\bibinfo{author}{Khanam, R.} \& \bibinfo{author}{Hussain, M.}
\newblock \bibinfo{title}{Yolov11: An overview of the key architectural enhancements}.
\newblock \bibinfo{howpublished}{{arXiv Preprint} arXiv:2410.17725}, \doiprefix\url{10.48550/arXiv.2410.17725} (\bibinfo{year}{2024}).

\end{thebibliography}



\section*{Author contributions statement}
B.C., J.L. and Y.Z. designed the study and performed the experiments and are the main contributing authors of the paper. B.C. contributed to the experiments and the figures for the paper. J.L. contributed to the image collection and the manuscript preparation. Z.X., M.P., X.M., Q.L and Q.M. contributed to the funding of the research. Y.Z. reviewed and approved the final manuscript.

\section*{Funding}
This research was funded by the Seed Precision Sorting Key Technology Integration and Equipment R\&D" project, funded by the 2024 Guangdong Provincial Rural Revitalization Strategy Special Funds (Document No. Yue Cai Nong [2024] No. 186)

\section*{Competing interests}
The authors declare no competing interests.

\section*{Data availability statement}
The data presented in this study are available on request from the corresponding author.

\section*{Additional information}
\textbf{Correspondence} and requests for materials should be addressed to Y.Z.


\end{document}